\documentclass{article}
\usepackage{times}
\usepackage{graphicx} 
\usepackage{natbib}
\usepackage{algorithm}
\usepackage{algorithmic}

\usepackage{color}
\usepackage{amsmath}
\usepackage{amsthm}
\usepackage{amssymb}
\usepackage{bm}

\newtheorem{thm}{Theorem}

\newtheorem{prop}[thm]{Proposition}

\usepackage{caption}

\input{Definitions}

\newcommand{\RN}[1]{%
	\textup{\lowercase\expandafter{\it \romannumeral#1}}%
}

\usepackage[accepted]{icml2017}

\icmltitlerunning{Stochastic Gradient Monomial Gamma Sampler}

\begin{document}

\twocolumn[
\icmltitle{Stochastic Gradient Monomial Gamma Sampler}




\begin{icmlauthorlist}
\icmlauthor{Yizhe Zhang}{duke}
\icmlauthor{Changyou Chen}{duke}
\icmlauthor{Zhe Gan}{duke}
\icmlauthor{Ricardo Henao}{duke}
\icmlauthor{Lawrence Carin}{duke}
\end{icmlauthorlist}

\icmlaffiliation{duke}{Duke University, Durham, NC, 27708}

\icmlcorrespondingauthor{Yizhe Zhang}{yizhe.zhang@duke.edu}

\icmlkeywords{MCMC, mixing rate}

\vskip 0.3in
]



\printAffiliationsAndNotice{}  

\begin{abstract}
	Recent advances in stochastic gradient techniques have made it possible to estimate posterior distributions from large datasets via Markov Chain Monte Carlo (MCMC).
	However, when the target posterior is multimodal, mixing performance is often poor.
	This results in inadequate exploration of the posterior distribution.
	A framework is proposed to improve the sampling efficiency of stochastic gradient MCMC, based on Hamiltonian Monte Carlo.
	A generalized kinetic function is leveraged, delivering superior stationary mixing, especially for multimodal distributions.
	Techniques are also discussed to overcome the practical issues introduced by this generalization.
	It is shown that the proposed approach is better at exploring complex multimodal posterior distributions, as demonstrated on multiple applications and in comparison with other stochastic gradient MCMC methods.
\end{abstract}
\section{Introduction}
The development of increasingly sophisticated Bayesian models in modern machine learning has accentuated the need for efficient generation of asymptotically exact samples from complex posterior distributions.
Markov Chain Monte Carlo (MCMC) is an important framework for drawing samples from a target density function.
MCMC sampling typically aims to estimate a desired expectation in terms of a collection of samples, avoiding the need to compute intractable integrals.
The Metropolis algorithm \citep{metropolis1953equation} was originally proposed to tackle this task.
Despite great success, this method is based on \emph{random walk} exploration, which often leads to inefficient posterior sampling (with a finite number of samples).
Alternatively, exploration of a target distribution can be guided using \emph{proposals} inspired by \emph{Hamiltonian dynamics}, leading to Hamiltonian Monte Carlo (HMC) \citep{duane1987hybrid}.
Aided by gradient information, HMC is able to move efficiently in parameter space, thus greatly improving exploration.
However, the emergence of big datasets poses a new challenge for HMC, as evaluation of gradients on whole datasets becomes computationally demanding, if not prohibitive, in many cases.

To scale HMC methods to big data, recent advances in Stochastic Gradient MCMC (SG-MCMC) have subsampled the dataset into \emph{minibatches} in each iteration, to decrease computational burden \citep{welling2011bayesian,chen2014stochastic,DingFBCSN:nips14,ma2015complete}.
Stochastic Gradient Langevin Dynamics (SGLD) \citep{welling2011bayesian} was first proposed to generate approximate samples from a posterior distribution using minibatches.
Since then, research has focused on leveraging the minibatch idea while also providing theoretical guarantees.
For instance, \citet{teh2014consistency} showed that by appropriately injecting noise while using a stepsize-decay scheme, SGLD is able to converge asymptotically to the desired posterior.
Stochastic Gradient Hamiltonian Monte Carlo (SGHMC) \citep{chen2014stochastic} extended SGLD with auxiliary momentum variables, akin to HMC, and introduced a \emph{friction} term to counteract the stochastic noise due to subsampling.
However, exact estimation of such noise is needed to guarantee a correct SGHMC sampler.
To alleviate this issue, the Stochastic Gradient Nos\'{e}-Hoover Thermostat (SGNHT) \citep{DingFBCSN:nips14} algorithm introduced so-called \emph{thermostat} variables to adaptively estimate stochastic noise via a thermal-equilibrium condition.

One standing challenge of SG-MCMC methods is inefficiency when exploring complex multimodal distributions.
This limitation is commonly found in latent variable models with a multi-layer structure.
Inefficiency is manifested because sampling algorithms have difficulties moving across modes, while traveling along the surface of the distribution.
As a result, it may take a very large number of iterations (posterior samples) to cover more than one mode, greatly limiting scalability.

We investigate strategies for improving mixing in SG-MCMC.
We propose the Stochastic Gradient Monomial Gamma Thermostat (SGMGT), building upon the Monomial Gamma Sampler (MGS) proposed by \citet{zhang2016monomial}.
They showed that a generalized kinetic function typically improves the stationary mixing efficiency of HMC, especially when the target distribution has multiple modes.
However, this advantage comes with numerical difficulties, and convergence problems due to poor initialization.
By defining a \emph{smooth} version of this generalized kinetic function, we can leverage its mixing efficiency, while satisfying the required conditions for stationarity of the corresponding stochastic process, as well as alleviating numerical difficulties arising from differentiability issues.
To ameliorate the convergence issues, we further introduce $i$) a sampler with an underlying elliptic stochastic differential equation system and $ii$) a resampling scheme for auxiliary variables (momentum and thermostats) with theoretical guarantees.
The result is an elegant framework to improve stationary mixing performance on existing SG-MCMC algorithms augmented with auxiliary variables.

\section{Preliminaries}
\paragraph{Hamiltonian Monte Carlo}
Suppose we are interested in sampling from a posterior distribution represented as $\pi(\theta|X) \propto p(X|\theta) p(\theta) = \exp[-U(\theta|X)]$, where $\theta$ denotes model parameters and $X = \{x_1, \ldots, x_N\}$ represents $N$ data points.
Assuming i.i.d. data, the \emph{potential energy function} $U(\theta|X)$ denotes the negative log posterior density, up to a normalizing constant, \emph{i.e.},
%
	$U(\theta|X) = - \sum_{i=1}^{N} \log p(x_{i}|\theta) - \log p(\theta)$. 
%
For simplicity, in the following we omit the conditioning of $X$ in $U(\theta|X)$, and write $U(\theta)$.
In HMC, the posterior density is augmented with an auxiliary momentum random variable $p$; $p$ is independent of $\theta$, and typically has a marginal Gaussian distribution with zero-mean and covariance matrix $M$.
The joint distribution is written as
%
	$p(\theta,p) \propto \exp[-H(\theta,p)] \ \triangleq \ \exp[-U(\theta)-K(p)]$, 
%
where $H(\theta,p)$ is the total energy (or \emph{Hamiltonian}) and $K(p) = \frac{1}{2} p^T M^{-1} p$ is the standard (Gaussian) \emph{kinetic energy function}, and $M$ is the mass matrix.
HMC leverages Hamiltonian dynamics, driven by the following differential equations
\begin{align}\label{eq:hamd}
	d \theta = M^{-1} p dt \,, \qquad d p = - \nabla U(\theta) dt \,,
\end{align}
where $t$ is the system's time index.

The total Hamiltonian is preserved under perfect simulation, \emph{i.e.}, by solving \eqref{eq:hamd} exactly.
However, closed-form solutions for $p$ and $\theta$ are often intractable, thus numerical integrators such as the \emph{leap-frog} method are utilized to generate approximate samples of $\theta$ \citep{neal2011mcmc}.
This leads to the following update scheme:
\begin{align}\label{eq:leapfrog}
	\begin{aligned}
		p_{t + 1 / 2} & = p_t - \tfrac{\epsilon}{2}\nabla U(\theta_t) \,, \\
		\theta_{t + 1} & = \theta_t + \epsilon  M^{-1} p_{t + 1 / 2} \,, \\
		p_{t + 1} & = p_{t + 1 / 2} - \tfrac{\epsilon}{2}\nabla U(\theta_{t + 1}) \,,
	\end{aligned}
\end{align}
where $\epsilon$ is the \emph{stepsize}.

\paragraph{Monomial Gamma HMC}
In the Monomial Gamma Hamiltonian Monte Carlo (MGHMC) \citep{zhang2016monomial} algorithm, the following \emph{generalized} kinetic function is employed as a substitute for the Gaussian kinetics of standard HMC:
\begin{align}\label{eq:km}
	K(p) = (|p|^{\frac {1} {2a}})^T M^{-1} |p|^{\frac {1} {2a}} \,,
\end{align}
where $|p|^{\frac {1} {2a}}$ denotes the element-wise power operation, $a$ is the monomial parameter.
Note that when $a=1/2$, \eqref{eq:km} recovers the standard (Gaussian) kinetics.
For general $a$, the update equations are identical to \eqref{eq:leapfrog}, except for
\begin{align}\label{eq:theta_k}
	\theta_{t + 1} & = \theta_t + \epsilon \nabla K (p_{t + 1 / 2}) \,.
\end{align}
\citet{zhang2016monomial} proved in the univariate case that MGHMC can yield better mixing performance when the sampler reaches its stationary distribution, under perfect dynamic simulation, \ie infinitesimal stepsize in the limit and adequate (finite) simulation stepsize.
Additionally, it was shown that for multimodal distributions sampled via MGMHC, the probability of getting trapped in a single mode goes to zero, as $a\to\infty$.

However, these theoretical advantages are accompanied by two practical issues: $i$) the numerical difficulties accentuate dramatically as $a$ increases, due to the lack of differentiability of $K(p)$ for $a \geq 1$, and $ii$) convergence is slow with poor initialization.
For example, in \eqref{eq:km} and \eqref{eq:theta_k}, if $\theta_{t}$ is far away from the mode(s) of the distribution, $\nabla U(\theta_t)$ will be large, causing the updated momentum $p_{t + 1 / 2}$ to blow up.
This renders the change of $\theta$, \ie $\nabla K (p_{t + 1 / 2})$, to be arbitrarily small for large $a$, thus slowing convergence.

\paragraph{Stochastic Gradient MCMC}
SG-MCMC is desirable when the dataset, $X$, is too large to evaluate the potential $U(\theta)$ using all $N$ samples.
The idea behind SG-MCMC is to replace $U(\theta)$ with an unbiased \emph{stochastic likelihood}, $\tilde{U} (\theta)$, evaluated from a subset of data (termed a minibatch)
\begin{align}\label{eq:stoc_g}
  \tilde{U} (\theta) = - \tfrac {N} {N'} \textstyle{\sum}_{i=1}^{N'} \log p(x_{\tau_i}|\theta) - \log p(\theta) \,,
\end{align}
where $\{\tau_1, \cdots, \tau_{N^\prime}\}$ is a random subset of $\{1, 2, \cdots, N\}$ of size $N^{\prime}\ll N$.
SG-MCMC algorithms are typically driven by a continuous-time Markov stochastic process of the form \citep{chen2015convergence}
\begin{align}\label{eq:MSP}
	d \Gamma  = V(\Gamma) dt + D(\Gamma) dW \,,
\end{align} 
where $\Gamma$ denotes the parameters of the \emph{augmented} system, \emph{e.g.}, $p$ and $\theta$, $V(\cdot)$ and $D(\cdot)$ are referred as \emph{drift} and \emph{diffusion} vectors, respectively, and $W$ denotes a standard Wiener process.

In SGHMC \citep{chen2014stochastic}, the resulting stochastic dynamic process is governed by the following Stochastic Differential Equations (SDEs) (with $M=I$):
\begin{align}\label{eq:sghmc}
	\begin{aligned}
		d \theta & = p {dt} \,, \\
		d p & = - [\nabla \tilde {U} (\theta) + A p] dt + \sqrt{2 (A I - \hat{B}(\theta)) } dW \,,
	\end{aligned}
\end{align}
where $\Gamma = \{\theta, p\}$, $V(\Gamma)$ is a function of $\{p,\nabla_\theta \tilde {U},A\}$, and $D(\Gamma)$ is a function of $\{A,\hat{B}(\theta)\}$. $\nabla \tilde{U}(\theta)$ is modeled as $\nabla \tilde{U}(\theta) =\nabla U(\theta)+\sqrt{2B(\theta)}\nu$, where $\nu\sim\mathcal{N}(0,1)$ and $h$ is the discretization stepsize.
$\hat{B}(\theta)$ is an estimator of $B(\theta)$, $A$ is a user-specified \emph{diffusion} factor and $I$ is the identity matrix.
\citet{chen2014stochastic} set $\hat{B}(\theta)=0$ for simplicity.
The reasoning is that the injected noise $\mathcal{N}(0,2Ah)$ will dominate as $h \rightarrow 0$ ($A$ remains as a constant), whereas $B(\theta)$ goes to zero.
Unfortunately, the covariance function, $B(\theta)$, of the stochastic noise, $\nu$, is difficult to estimate in practice.

Recently, SGNHT \citep{DingFBCSN:nips14} considered incorporating additional auxiliary variables (thermostats). 
The resulting SDEs correspond to
\begin{align}\label{eq:sgnht}
		d p & = - [\nabla  \tilde{U} (\theta) + \xi \odot p] dt +  \sqrt{2 A I  } dW \,, \\
		d \theta & = p {dt}\, , \, d \xi  = (p \odot p -1)dt \,,
\end{align}
where $\odot$ represents the Hadamard (element-wise) product and $\xi$ are thermostat variables.
Note that the diffusion factor, $A$, is decoupled in \eqref{eq:sgnht}, thus $\xi$ can adaptively fit to the unknown noise from the stochastic gradient $\nabla \tilde{U} (\theta)$.

\section{Stochastic Gradient Monomial Gamma Sampler}
We now consider $i$) a more efficient (generalized) kinetic function, $ii$) adapting the proposed kinetics to satisfy stationary requirements and alleviate numerical difficulties, $iii$) incorporating an additional first-order stochastic process to \eqref{eq:sgnht} and $iv$) stochastic resampling of the momentum and thermostats to lessen convergence issues.

\paragraph{Generalized kinetics}
The statistical physics literature traditionally considers a quadratic form of the kinetics function, and a Gaussian distribution for the thermostats in \eqref{eq:sgnht}, when analyzing the dynamic system of a canonical ensemble \citep{tuckerman2010statistical}.
Inspired by this, one typical assumption in previous SG-MCMC work is that the marginal distribution for the momentum and thermostat is Gaussian \citep{DingFBCSN:nips14,li2015high}.
However, this assumption, while convenient, does not necessarily guarantee an optimal sampler.

In recent work, \citet{lu2016relativistic} extended the standard (Newtonian) kinetics to a more general form inspired by relativity theory.
By bounding the momentum, their \emph{relativistic Monte Carlo} can lessen the problem associated with large potential gradients, $\nabla U(\theta_t)$, thus resulting in a more robust alternative to standard HMC.
Further, \citet{zhang2016monomial} demonstrated that adopting non-Gaussian kinetics delivers better mixing and reduces sampling autocorrelation, especially for cases where the posterior distribution has multiple modes.

These ideas motivate a more general framework to characterize SG-MCMC, with potentially non-Gaussian kinetics and thermostats.
As a relaxation of SGNHT \citep{DingFBCSN:nips14, ma2015complete}, we consider a Hamiltonian system defined in a more general form
\begin{align}\label{eq:Hamilton}
	H = K (p) + U (\theta) + F (\xi) \,,
\end{align}
where $K(\cdot)$ and $F(\cdot)$ are any valid potential functions, inherently implying that $\exp[-K(\cdot)]$ and $\exp[-F(\cdot)]$, define valid probability density functions.

We first consider the SDEs of SGNHT with generalized kinetics $K(p)$.
The system can be obtained by generalizing $K(p) = p^T p/2 $ (with identity mass matrix $M$ for simplicity) in \eqref{eq:sgnht} with arbitrary $K(p)$, thus
\begin{align}\label{eq:sgmgt}
	\begin{aligned}
		d \theta & = \nabla K(p) {dt} \,, \\
		d p & = - [\nabla  \tilde{U} (\theta) + \xi \odot \nabla K(p)] dt +  \sqrt{2 A I  } dW \,, \\
		d \xi & = (\nabla K(p) \odot \nabla K(p) -\nabla^2 K(p))dt \,.
	\end{aligned}
\end{align}
However, if we set $K(p)$ as in \eqref{eq:km} with $a \geq 1$, the dynamics governing the SDEs in \eqref{eq:sgmgt} will often fail to converge.
This is because the sufficient condition to guarantee that the It\^{o} process governed by the SDEs in \eqref{eq:sgmgt} converge to a stationary distribution generally requires the Fokker-Planck equation to hold \citep{risken1984fokker}.
Further, the existence and uniqueness of the solutions to the Fokker-Planck equation require Lipschitz continuity of drift and diffusion vectors in \eqref{eq:MSP} \citep{bris2008existence}.
Unfortunately, this is not the case for the drift vectors in \eqref{eq:sgmgt} when $a \geq 1$, as $\nabla K(p)$ is non-differentiable at the origin, \emph{i.e.}, $p=0$.

\begin{figure}[t]
	\centering
	\includegraphics[width=.18\textwidth]{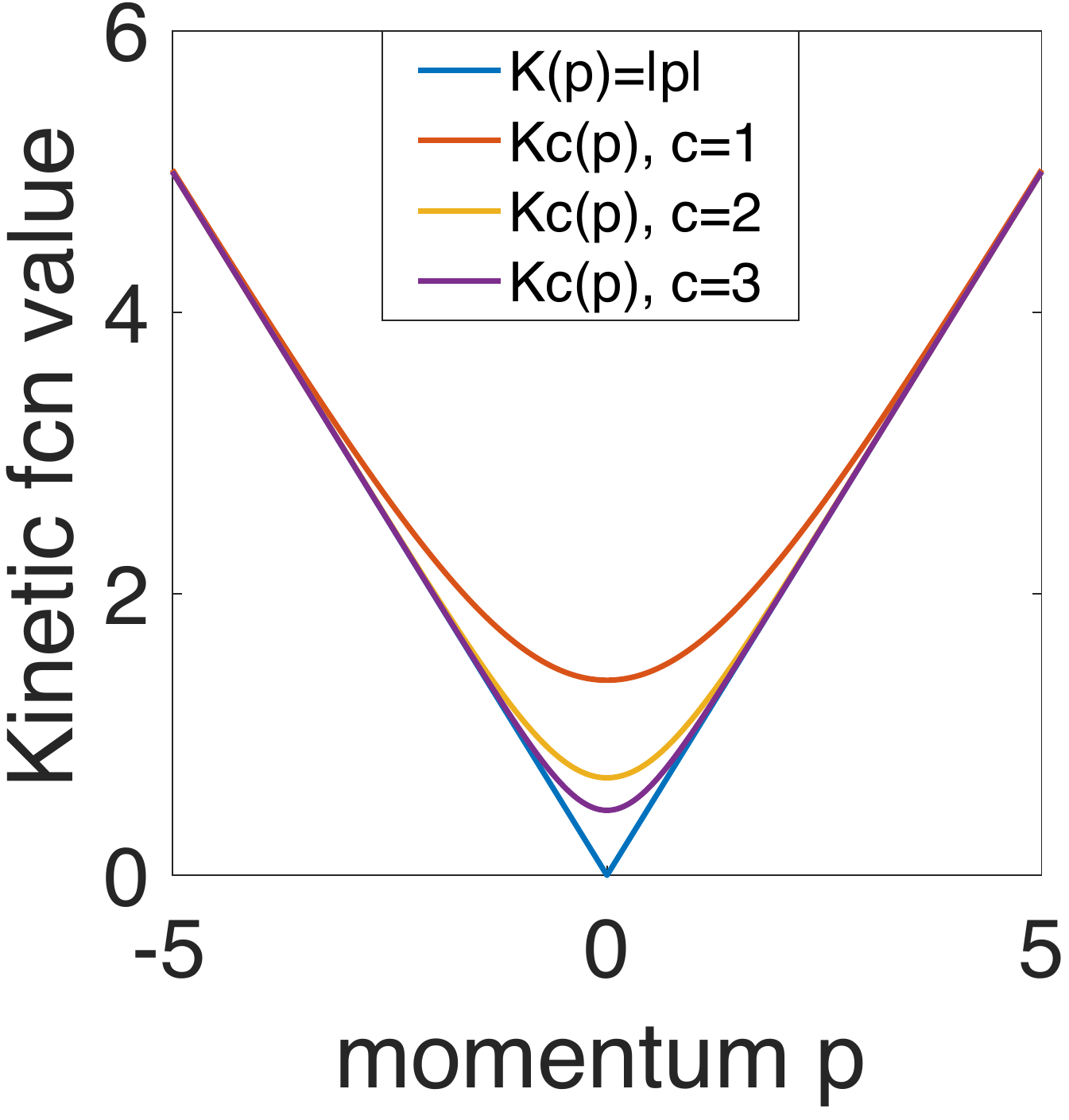}\hspace{8mm}
	\includegraphics[width=.18\textwidth]{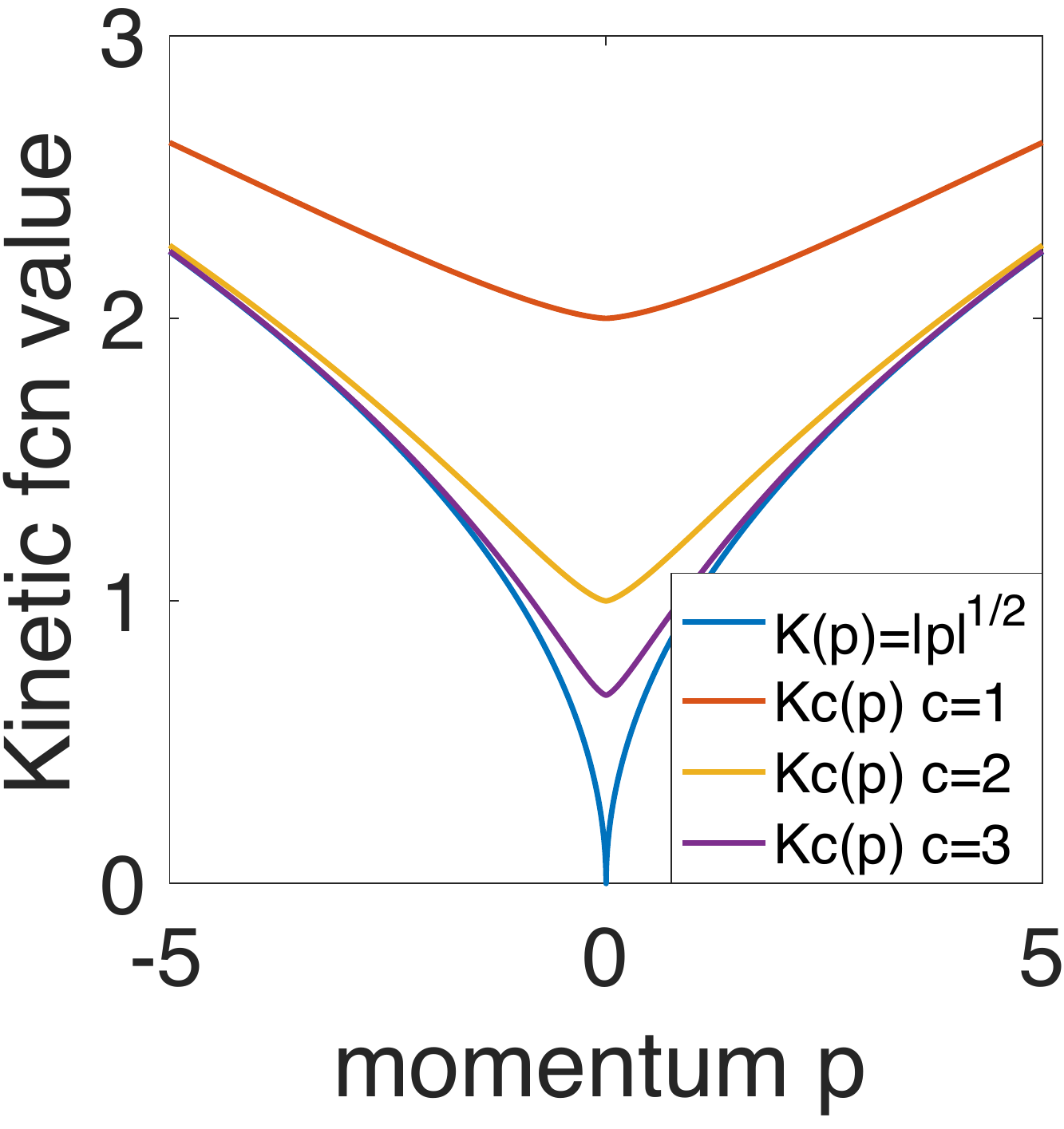}
	\caption{Softened \emph{vs.} stiff kinetics (1D). Left: $a=1$. Right: $a=2$.} \label{fig:soft}
	\vspace{-2mm}
\end{figure}

\paragraph{Softened kinetics}
The above limitation can be avoided by using a \emph{softened} kinetic function $K_c(p)$.
However, to keep the performance benefits from the original \emph{stiff} kinetics, we must ensure that $K_c(p)$ has the same tail behavior.
We propose that for $a=\{1,2\}$, the softened kinetics are (for clarity we consider 1D case, however higher dimensions still apply)
\begin{align}\label{eq:soften_new}
	K_c(p) = \left\{\begin{array}{lr}
	        - p + 2/c \log (1 + e^{c p}), & a=1 \\
	        | p |^{1/2} + \frac {4}  {c(1 + e^{c | p |^{1/2}})}, & a=2
	        \end{array}\right. \,,
\end{align}
where $c>0$ is a \emph{softening} parameter.
Note that $K_c(p)$ is (infinitely) differentiable for any $c$ and asymptotically approaches the stiff kinetics as $c \rightarrow \infty$.
A comparison between stiff kinetics, $K(p)$, and softened kinetics $K_c(p)$ is shown in Figure~\ref{fig:soft}, for different values of $c$.
Discussion and formulation of the softened kinetics for arbitrary $a$ (and $M$) are provided in the Supplementary Material (SM).

To generate samples of the momentum variable, $p$, from the density with softened kinetics, which is proportional to $\exp[-K_c(p)]$, we use a coordinate-wise rejection sampling, \emph{i.e.}, the proposed $p_d$ for the $d$-th dimension is rejected with probability $1-\exp[K(p_d)-K_c(p_d)]$.

In practice, setting $c$ to a relatively large value would still make the gradient $\nabla K_c(p)$ ill-posed close to $p=0$, thus causing high \emph{integration error} when simulating the Hamiltonian dynamics.
Conversely, setting $c$ to a small value will cause a \emph{high approximation} error w.r.t. the original $K(p)$, thus resulting in a less efficient sampler.
Consequently, $c$ has to be determined empirically as a trade-off between integration and approximation errors.

\paragraph{Additional First Order Dynamics}
Inspired by \citet{ma2015complete}, we consider adding Brownian motion to $\theta$ and $\xi$ in \eqref{eq:sgnht}, with variances $\sigma_\theta$ and $\sigma_\xi$, respectively, while maintaining the stochastic process (asymptotically) converging to the correct marginal distribution of $\theta$.
Specifically, we consider the following SDEs:
\begin{align}\label{eq:generalized_sde}
	\begin{aligned}
		d \theta = & - \sigma_{\theta} \nabla \tilde{U} (\theta) dt + \nabla K_c (p) dt + \sqrt{2\sigma_{\theta} } dW \,, \\
		d p = & - (\sigma_p + \gamma \nabla F (\xi) ) \odot \nabla K_c (p) dt \\
		    & - \nabla \tilde{U} (\theta) dt + \sqrt{2\sigma_{p} } dW \,, \\
		d \xi = & \ \gamma \left[\nabla K_c (p)\odot \nabla K_c (p)  - \nabla^2 K_c (p)\right] dt \\
			& - \sigma_{\xi} \nabla F (\xi) dt + \sqrt{2\sigma_{\xi}} dW \,.
	\end{aligned}
\end{align}
The variances $\{\sigma_{\theta},\sigma_p,\sigma_{\xi} \}$ control the Brownian motion for $\{\theta,p,\xi\}$, respectively, and $\gamma>0$ denotes a rescaling factor for the friction term of momentum updates.
The additional terms $-\sigma_{\theta} \nabla \tilde{U} (\theta) dt  + \sqrt{2\sigma_{\theta} } dW $ and $- \sigma_{\xi} \nabla F (\xi) dt + \sqrt{2\sigma_{\xi}} dW$ can be understood as first-order Langevin dynamics \citep{welling2011bayesian}.
The variance term, $\sigma_{\theta}$, controls the contribution of $\nabla \tilde{U}(\theta)$ to the update of $\theta$ w.r.t. $\nabla K_c(p)$.
This is analogous to the hyperparameter balancing $\nabla \tilde{U}(\theta)$ and $p$ in the SGD-with-momentum algorithm \citep{rumelhart1988learning}.
Derivation details for $\nabla K_c (p)$ and $\nabla^2 K_c (p)$ in \eqref{eq:soften_new}, as well as other values of $a$, are provided in the SM.

The following theorem, proven in the SM, shows that under regularity conditions, the SDEs in \eqref{eq:generalized_sde} lead to posterior samples from the invariant joint distribution $p(\Gamma) \propto \exp[-H(\Gamma)]$, yielding the desired marginal distribution w.r.t. $\theta$ as $p(\theta) \propto \exp[-U(\theta)]$.
\begin{thm}
	The stochastic process governed by \eqref{eq:generalized_sde} converges to a stationary distribution $p(\Gamma) \propto \exp[-H(\Gamma)]$, where $H(\Gamma)$ is as defined in \eqref{eq:Hamilton}, and $\Gamma=\{\theta,p,\xi\}$.
\end{thm}

The reasoning behind increasing \emph{stochasticity} in the SDEs is two-fold.
First, the additional Langevin dynamics are crucial to SG-MCMC with generalized kinetics for large $a$.
For instance, for $\sigma_{\theta} = 0$, the update for $\theta$ from \eqref{eq:sgmgt} is $\theta_{t+1} = \theta_{t} + \nabla K(p_t)h$.
When $a>1$ and $|p_t|$ is large, $\nabla K(p) = \frac{1}{a}|p|^{1/a-1}$ will be close to zero, thus $\theta_{t+1}$ (the next sample) will be close to $\theta_t$, \emph{i.e.}, the sampler moves arbitrarily slow.
As discussed by \citet{zhang2016monomial}, this can happen when $\theta$ moves to a region where the gradient $\nabla U(\theta)$ takes a large absolute value, \emph{e.g.}, near the low-density regions in a light-tailed distribution.
Fortunately, the additional Langevin dynamics in \eqref{eq:generalized_sde}, $-\sigma_{\theta} \nabla \tilde {U} (\theta) dt + \sqrt{2 \sigma_{\theta}} dW$, compensate for the weak updating signal from $\nabla K(p)$, by an immediate gradient signal $\nabla \tilde {U} (\theta)$.
Additionally, when $\tilde {U} (\theta)$ becomes small, $\nabla K(p)$ will become large.
As a result, these two updating signals $\nabla K(p)$ and $\nabla \tilde {U} (\theta)$ compensate each other, thereby delivering a stable updating scheme.
Likewise, the immediate gradient $\nabla F (\xi)$ in \eqref{eq:generalized_sde} can provide complementary updating signal for the thermostat variables, $\xi$, to offset the weak deterministic update $\nabla K_c (p)\odot \nabla K_c (p)  - \nabla^2 K_c (p)$, when $p$ is large.

Second, \eqref{eq:generalized_sde} has noise components on all parameters, $\{\theta,p,\xi\}$, making the corresponding SDEs {\it elliptic}.
From a theoretical perspective, ellipticity/hypoellipticity are necessary conditions to guarantee existence of bounded solutions for a particular partial differential equation related to the diffusion's {\em infinitesimal generator}, which lies in the core of most recent SG-MCMC theory \citep{teh2014consistency,vollmer2016exploration,chen2015convergence}. Ellipticity is characterized by a noise process (Brownian motion) covering all components of the system, via the diffusion, $D(\Gamma)$, in \eqref{eq:MSP}.
This means $D(\Gamma)$ is block diagonal, thus a positive definite matrix \citep{MattinglyST:JNA10}.
In a typical hypoelliptic case, the noise process is imposed on a subset of $\Gamma$.
However, hypoellipticity also requires the noise to be able to spread through the system via the drift term, $V(\Gamma)$, which may not be true for general $V(\Gamma)$. For instance, in \eqref{eq:sgnht}, $\Gamma=\{\theta,p,\xi\}$ and $D(\Gamma)$ is block diagonal with entries $\{0,\sqrt{2AI},0\}$, \emph{i.e.}, $\theta$ and $\xi$ are not explicitly influenced by the noise process, $W$, thus hypoellipticity cannot be guaranteed.

To the authors' knowledge, for existing SG-MCMC algorithms, only SGLD where $d\theta=-\nabla_\theta\tilde{U}(\theta)dt+\sqrt{2}dW$, satisfies the ellipticity property, while other algorithms such as SGHMC and SGNHT assume hypoellipticity, thus their corresponding $D(\Gamma)$ are not positive definite.
%

One caveat of \eqref{eq:generalized_sde} is that if $\sigma_{\theta}$ and $\sigma_{\xi}$ are too large, the updates will be dominated by first-order dynamics, thus losing the convergence benefits from second-order dynamics \cite{chen2014stochastic}.
In practice, $\sigma_{\theta}$ and $\sigma_{\xi}$ are problem-specific, thus need to be tuned, \eg, by cross-validation.

\paragraph{Stochastic resampling}
When generating samples from the stochastic process in \eqref{eq:generalized_sde}, we resample momentum and thermostats from their marginal distribution with a fixed frequency, instead of every iteration from their conditionals.
Since the momentum and thermostats are drawn from the independent marginals of stationary distribution $p(\Gamma) \propto \exp[-H(\Gamma)]$, it can be shown that reconstructing the stochastic process with the solution of the SDEs will still leave the stochastic process invariant to the target stationary distribution \cite{brunick2013mimicking}.

\begin{figure}[t!]
	\centering
	\includegraphics[width=.37\textwidth]{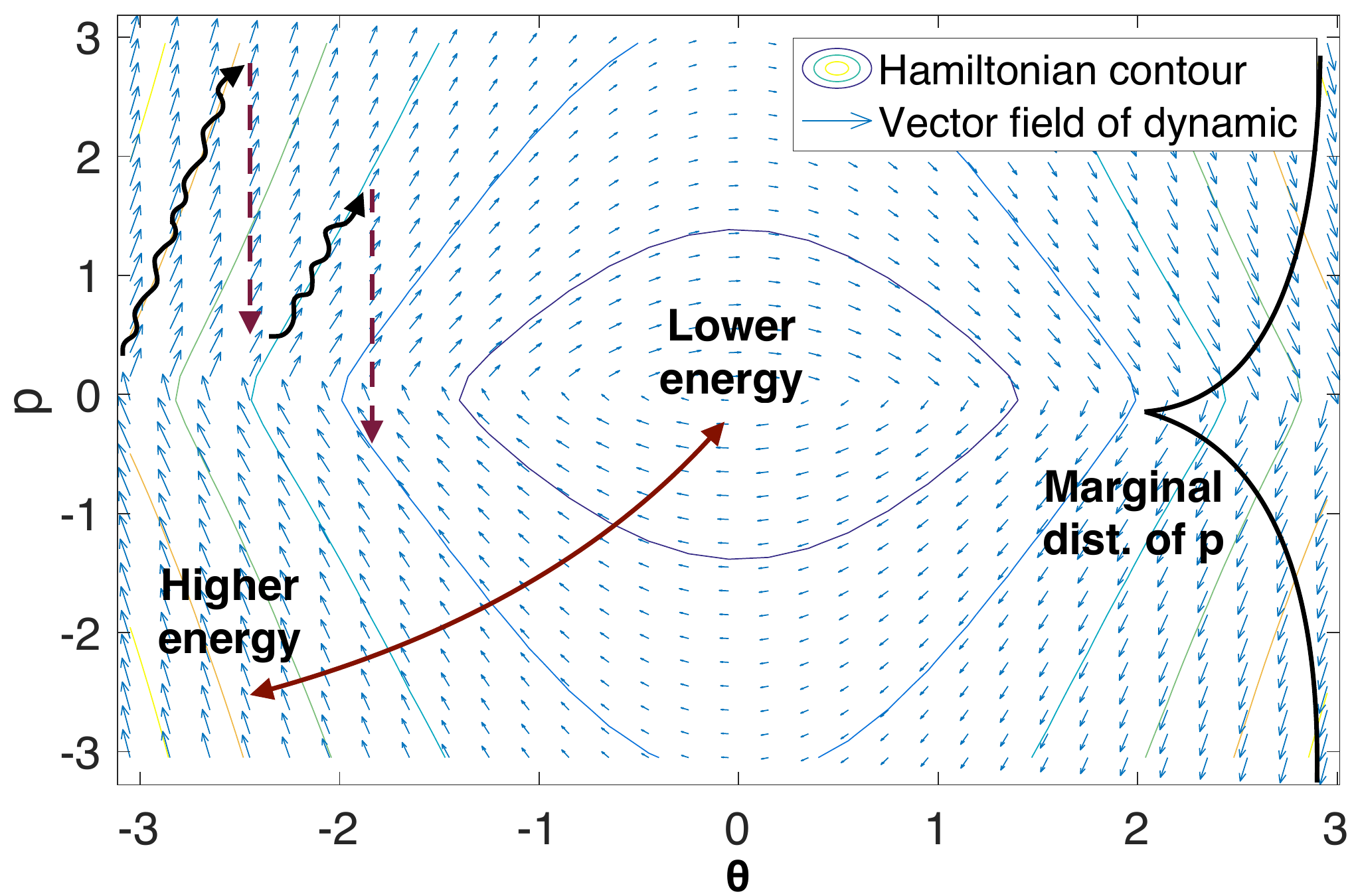}
	\includegraphics[width=.38\textwidth]{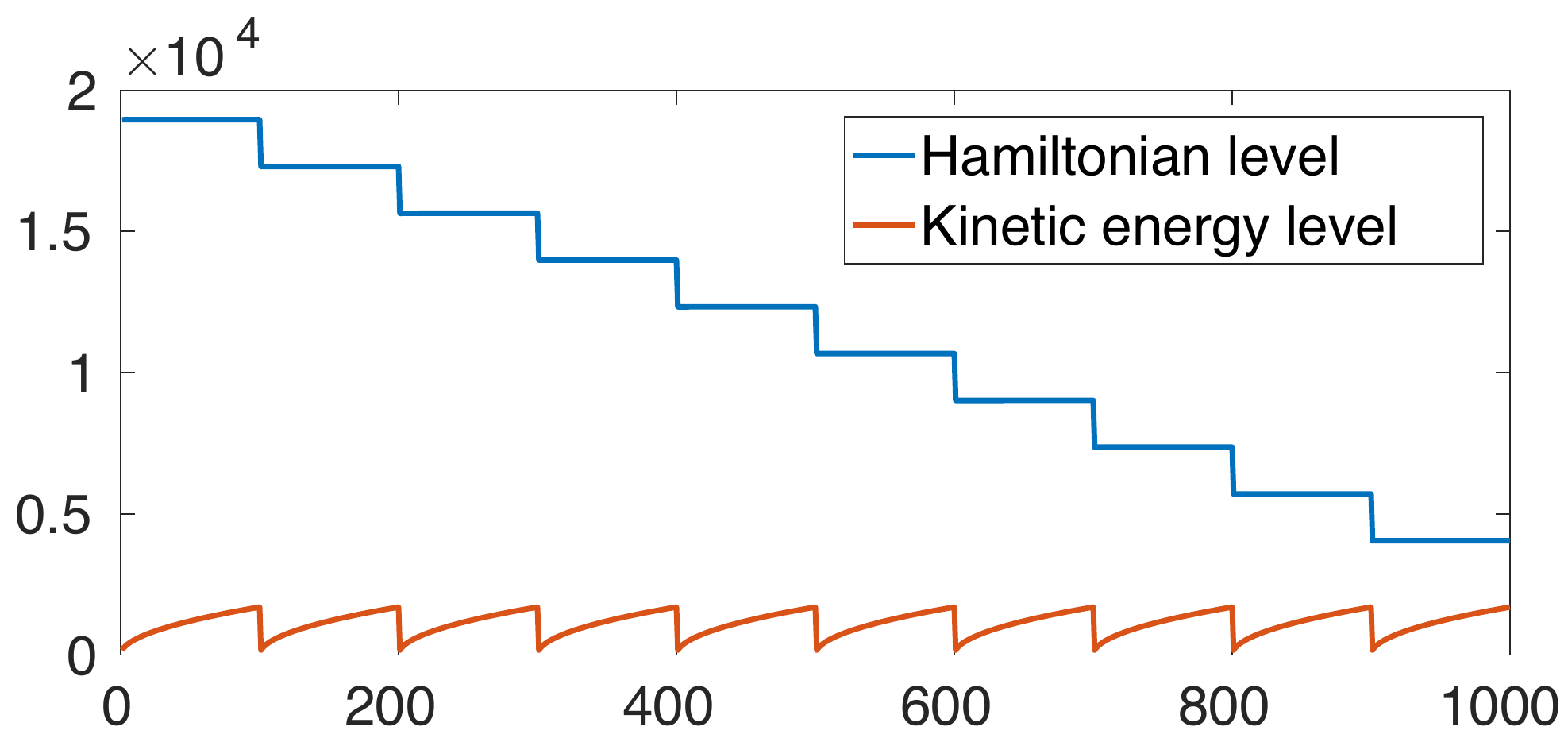}
	\caption{Momentum resampling. Top: stochastic process with resampling helps sampler move quickly to a lower Hamiltonian contour. Bottom: resampling decreases energy step-wise during burn-in stage. Resampling of $p$ occurs every 100 iterations.}
	\label{fig:dropping_energy}
	\vspace{-2mm}
\end{figure}

To simplify the discussion, consider a stochastic process of a particle $\{\theta,p\}$ as in \eqref{eq:sgmgt} with fixed $\xi$.
As show in Figure~\ref{fig:dropping_energy}, suppose the initial value of $\theta$ is far from the maximum \emph{a posteriori} value.
The dynamics governed by \eqref{eq:sgmgt} will stochastically move along the Hamiltonian contour.
The total Hamiltonian energy level is affected by the joint effect of the stochastic diffusion and momentum refraction (\emph{i.e.}, -$\xi pdt$), which changes continuously over time.

From previous discussions, moving on a high Hamiltonian contour when $a>1$ is less efficient because the absolute value of the momentum, $|p|$, will get increasingly large, slowing down the movement of $\theta$.
Resampling of momentum according to its marginal will enable the sampler to immediately move to a lower Hamiltonian energy level.

At the burn-in stage, this \emph{momentum-accumulation/energy-drop} cycle seen in Figure \ref{fig:dropping_energy}(bottom) via resampling momentum can happen several times, until equilibrium is found.
In practice, the resulting energy level is often much lower than initially, thereby delivering a more efficient and accurate dynamic updating.

The frequency of resampling from the marginal of the stationary distribution can have a direct impact on the mixing performance.
Setting the frequency too high will result in a random-walk behavior.
Conversely, with a low frequency resampling, the random-walk behavior is suppressed at a cost of fewer jumps between trajectories associated with different energy levels. It is advisable to increase the resampling frequency if the sampler is initialized on low-density (\eg light-tailed) region.

The resampling step on $p$ and $\xi$ plays a role that is similar to adding a Langevin component to $\theta$, in the sense that both improve convergence for $a> 1$.
However, these two strategies (resampling and Langevin) are fundamentally different.
We empirically observe that resampling is most helpful during burn-in, while the additional Langevin-style updates are more helpful with mixing during stationary sampling.

\paragraph{SGMGT}
The specifications described above constitute an SG-MCMC method for the SDEs in \eqref{eq:sgmgt}, which we call Stochastic Gradient Monomial Gamma Thermostat (SGMGT). 
We denote the SG-MCMC method with additional Brownian motion on $\theta$ and $\xi$ in \eqref{eq:generalized_sde} as SGMGT-D (Diagonal), \emph{i.e.}, with $\sigma_{\theta}>0$ and $\sigma_{\xi}>0$.
The complete update scheme, with Euler integrator, for SGMGT is presented in the SM.
Note that with $a=1/2, \sigma_{\theta} = 0, \sigma_{\xi} \rightarrow 0, c \rightarrow \infty$, SGMGT-D recovers SGHMC as in \citet{chen2014stochastic}. 
Moreover, when $a=1/2, \sigma_{\theta} = 0, c \rightarrow \infty$, it becomes SGNHT as in \citet{DingFBCSN:nips14}.

We note that SGMGT-D improves upon SGNHT in three respects:
($i$) we introduce generalized kinetics, which provably yield lower autocorrelations than standard HMC, especially in multimodal cases;
($ii$) the additional stochastic noise on thermostat variables yields more efficient mixing;
($iii$) we use stochastic resampling to allow for faster interchange between different energy levels, thus alleviating sampling \emph{stickiness}.

To the authors' knowledge, despite existing analysis for Langevin Monte Carlo \cite{bubeck2015finite,dalalyan2016theoretical}, rigorous analysis and comparison of the mixing performance of general SG-MCMC is very difficult, thus not yet established. 
Toward understanding the mixing performance of SGMGT-D, we argue that as the minibatch size increases, and the contribution of the diffusion in \eqref{eq:MSP} decreases, the SGMGT-D will approach MGHMC, in which case, a large $a$ will result in high stationary mixing performance, especially when sampling multimodal distribution, as theoretically shown by \citet{zhang2016monomial}.
Although our experiments support our intuition, a more formal theoretical justification is needed.
We leave this as interesting future work.

We observe empirically that when increasing the value of $a$, SGMGT-D may not always achieve superior mixing performance.
One possible reason for this is a larger value of $a$ induces ``stiffer'' behavior of $\exp[-K(p)]$ at $p=0$, which typically requires a higher level of softening, thus higher rejection rates during the rejection sampling step.
Also, when the dimensionality of $p$ is higher, the rejection rate of the rejection sampling step increases (proportional to $p$).
In such cases, the efficiency of the sampler decreases with large $a$.
For these reasons, we limit our experiments to $a=\{1,2\}$.

We clearly have more hyperparameters than SGNHT.
In practice, we fix $M=I$, $a=\{1,2\}$, and set the resampling frequency $T_p=T_{\xi}=100$, which provides robust performance.
Thus, only two additional hyperparameters are employed ($\sigma_{\theta}$ and $\sigma_{\xi}$) compared to SGNHT, and these parameters require further tuning.
We use either validation or a hold-out set in our experiments.

\paragraph{More accurate numerical integrators}
Using a first-order Euler integrator to approximate the solution of the continuous-time SDEs in \eqref{eq:generalized_sde}, leads to $O(h)$ errors in the approximate samples \citep{chen2015bridging}.
Alternatively, we can use the symmetric splitting scheme of \citet{chen2015bridging} to reduce the order of the approximate error to $O(h^2)$.
Details of the splitting used in this work are provided in the SM.

\paragraph{Convergence properties}
The SGMGT framework, as an instance of SG-MCMC, enjoys the same convergence properties of general SG-MCMC algorithms studied in \citet{chen2015convergence}. It's worth to mention that on challenging problems the posterior may not be densely sampled to yield ideal posterior computation, and the asymptotic theory is being used as a useful heuristic.
Specifically, it is of interest to quantify how fast the sample average, $\hat{\phi}_T$, converges to the true posterior average, $\bar{\phi} \triangleq \int \phi(\theta) \pi(\theta | X) \mathrm{d}\theta$, for $\hat{\phi}_T \triangleq \frac{1}{T}\sum_{t=1}^{T}\phi(\theta_t)$, where $T$ is number of iterations. 
Here we make the same assumptions of \citet{chen2015convergence}, and further assume that a first-order Euler integrator and a fixed stepsize are used. 
\begin{prop}\label{prop:biasmse}
	For the proposed SGMGT and SGMGT-D algorithms, if a fixed stepsize $h$ is used, we have:
	\begin{align*}
		\mbox{Bias: }&\left|\mathbb{E}\hat{\phi}_T - \bar{\phi}\right| = O\left(1/(Th) + h\right) \,, \\
		\mbox{MSE: }&\mathbb{E}\left(\hat{\phi} - \bar{\phi}\right)^2 = O \left(1/(Th) + h^{2}\right) \,.
	\end{align*}
\end{prop}

This proposition indicates that with larger number of iterations and smaller step sizes, smaller bias and MSE bounds can be achieved.
We note that these bounds have similar rates compared to other SG-MCMC algorithms such as SGLD, however, as we demonstrate below in the experiments, SGMGT and SGMGT-D usually converge faster than existing SG-MCMC methods.

In addition, for stochastic resampling, we can extend Proposition 2 to the following complementary results:

\begin{lemma}
	Let $\pi_h$ be the stationary distribution of SGMGT-D. The stationary distribution of SGMGT-D with momentum resampling is the same as $\pi_h$.
\end{lemma}

\begin{lemma}
	The optimal finite-time bias and MSE bounds for SGMGT-D with momentum replacement remain the same as SGMGT-D.
\end{lemma}

Proofs of Lemma 1 and Lemma 2 are provided in the SM.
The proposed SGMGT framework has a strong connection with second-order stochastic optimization methods, leading to a sampling scheme with minibatches with similar mixing performance as slice sampling \cite{neal2003slice}.
We discuss the details of this in the SM.

\begin{figure}[t!]
	\begin{center}
	\includegraphics[width=.19\textwidth]{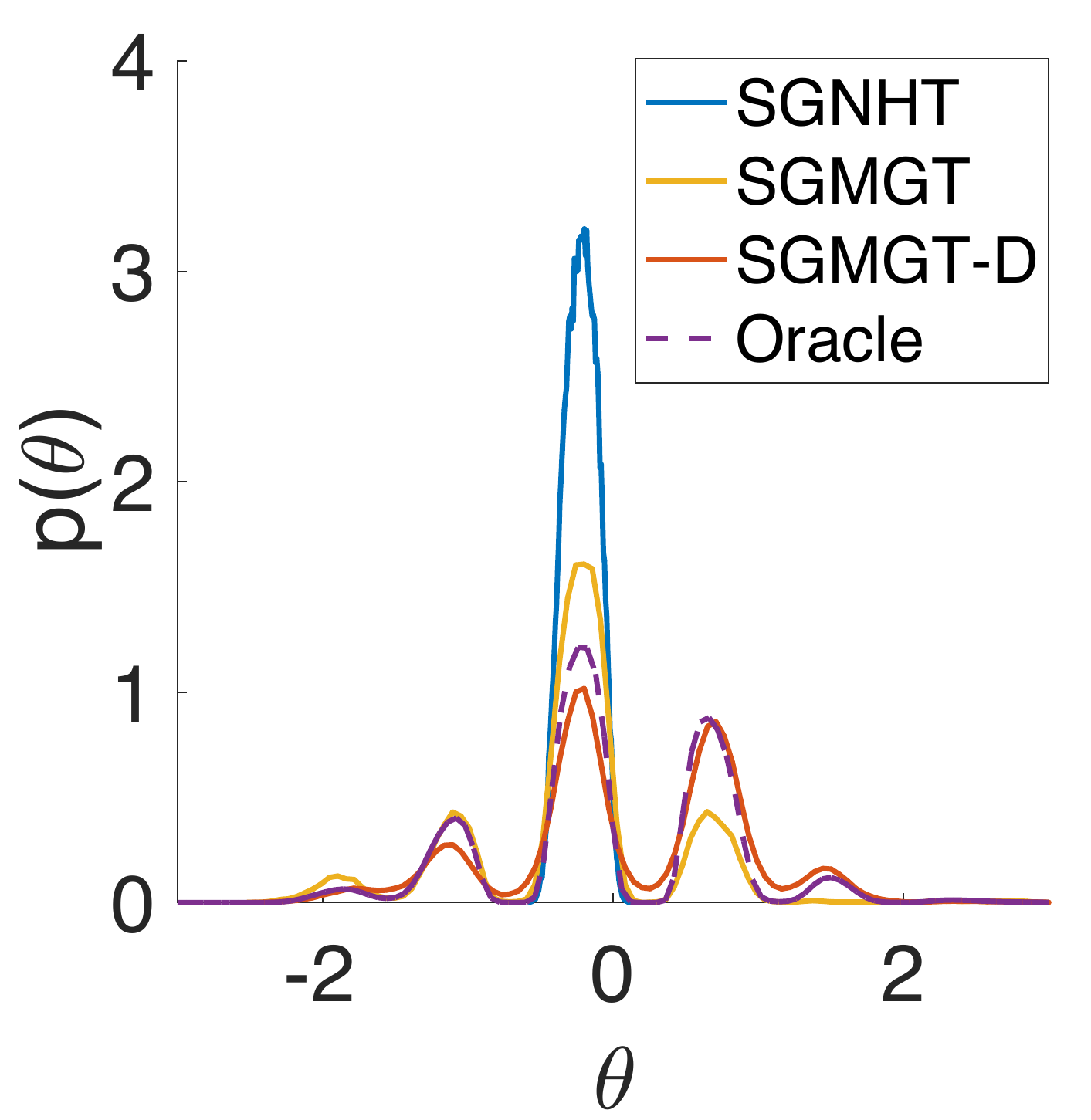}\hspace{4mm}
    \includegraphics[width=.21\textwidth]{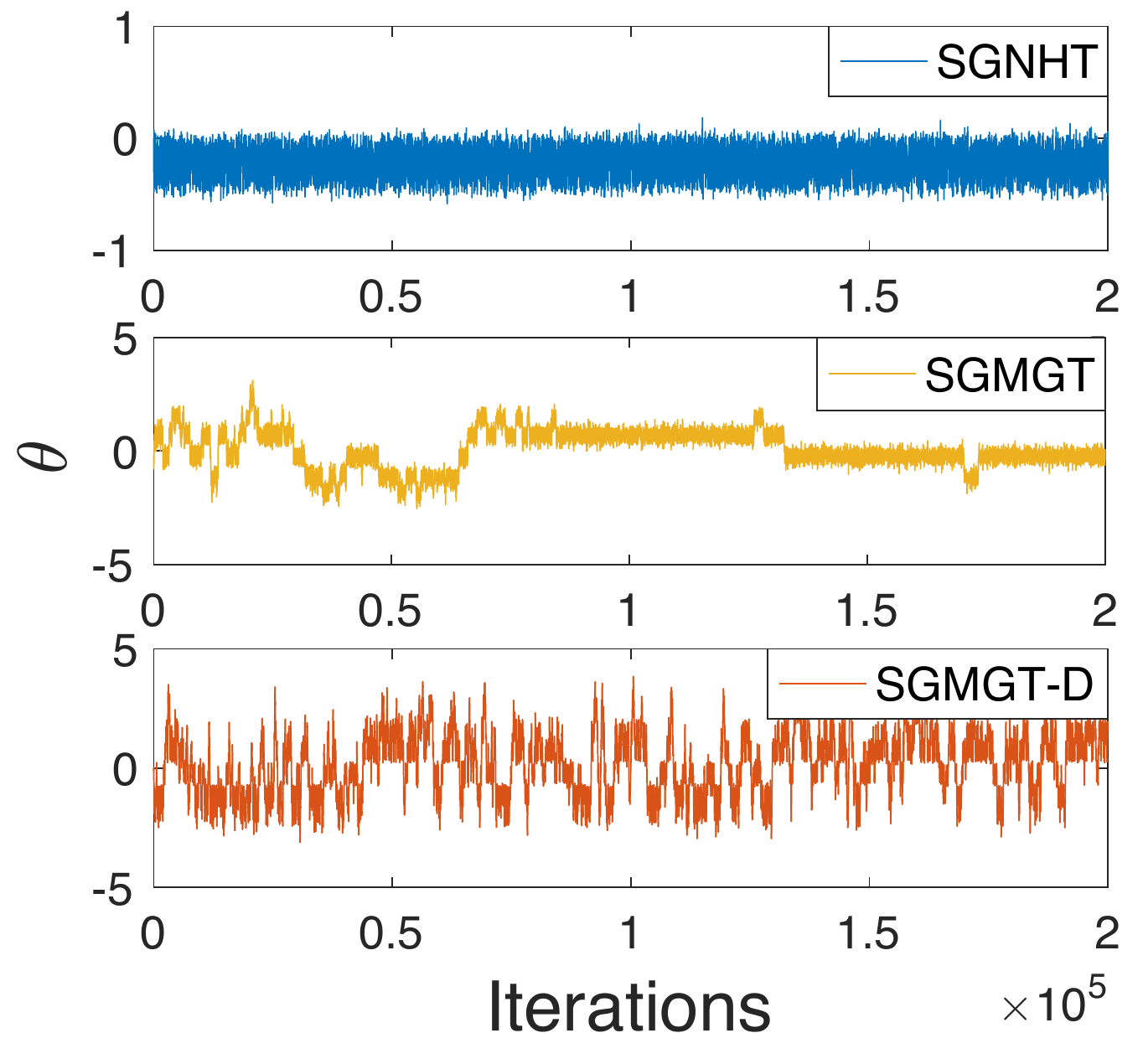}
	\end{center}
	\vspace{-4mm}
	\caption{Synthetic multimodal distribution. Left: empirical distributions for different methods. Right: traceplot for each method.}
	\label{fig:fig_fakir}
	\vspace{-3mm}
\end{figure}
\section{Experiments}
%
\subsection{Multiple-well Synthetic Potential}
We first evaluate the mixing efficiency of SGMGT and SGMGT-D for a synthetic problem, to generate samples from a complex multimodal distribution.
The distribution is shown in Figure~\ref{fig:fig_fakir}(left).
See SM for the definition of its potential energy.
The modes are almost isolated with a low-density region connecting each other.
Consequently, traversing between modes is difficult.
In order to simulate the noise of the gradient estimates, we set $\nabla \tilde{U}(\theta) = \nabla U(\theta) + \mathcal{N}(0,2B)$, similar to \citet{DingFBCSN:nips14}, where $B=1$.

We compare SGNHT with SGMGT and SGMGT-D with monomial parameter $a=2$ and fix $\gamma = 1$.
For all three algorithms, we try a number of hyperparameter settings, {\it e.g.}, stepsize $h$, $\{\sigma_{\theta}, \sigma_p, \sigma_{\xi}\}$, and the soft parameter $c$, and present the best results in Figure~\ref{fig:fig_fakir}.
Standard SGNHT fails to escape from one of the modes of the distribution.
For SGMGT with $a=2$, the generated samples reached 3 modes.
For SGMGT-D with $a=2$, the sampler identified all 5 modes.
In Figure~\ref{fig:fig_fakir}(right), SGMGT-D adequately moves across different modes and yields rapid mixing performance, unlike SGMGT which exhibits stickier behavior.

\begin{table}[t!]
	\caption{Average AUROC and median ESS. Dataset dimensionality is indicated in parenthesis after the name of each dataset.}
	\label{tab:BLR_acc}	
	\vskip 0.05in
	\begin{center}\scriptsize
		\begin{tabular}{ccccccc}
			AUROC ($D$) & A (15) & G (25) & H (14) & P(8) &
			R (7) & C (87)\\
			\hline
			SGNHT & 0.89 & 0.75 & 0.90 & 0.86 & 0.95 & 0.65 \\
			SGMGT(a=1) & 0.92 & 0.78 & 0.91 & 0.86 & 0.87 & 0.70\\
			SGMGT-D(a=1) & 0.95 & 0.86 & 0.95 & {\bf 0.93} & {\bf 0.98} & {\bf 0.73} \\
			SGMGT(a=2) & 0.93 & 0.79 & 0.93 & 0.88 & 0.86 & 0.62\\
			SGMGT-D(a=2) & {\bf 0.95} & {\bf 0.90} &  {\bf 0.95} & 0.90 & 0.97 & 0.69 \\
			\hline
			ESS ($D$) & A (15) & G (25) & H (14) & P(8) &
			R (7) & C (87)\\
			\hline
			SGNHT & 869 & 941 & 1911 & 2077 & 1761 & 1873 \\
			SGMGT-D(a=1) & {\bf 3147} & {\bf 2131} & 2448 & {\bf 4244} & 1494 & {\bf 3605} \\
			SGMGT-D(a=2) & 2700 & 1989 & {\bf 2768} & 3430 &  {\bf 2265} & 2969 \\
		\end{tabular}
	\end{center}
	\vspace{-6mm}
\end{table}
\subsection{Bayesian Logistic Regression}
We evaluated the mixing efficiency and accuracy of SGMGT and SGMGT-D using Bayesian logistic regression (BLR) on 6 real-world datasets from the UCI repository \citep{bache2013uci}: German
credit (G), Australian credit (A), Pima Indian (P), Heart (H), Ripley (R) and Caravan (C).
The data dimensionality ranges from 7 to 87, and total observations vary between 250 to 5822.
Gaussian priors are imposed on the regression coefficients.
We set the minibatch size to 16.
Other hyperparameters are provided in the SM.
For each experiment, we draw 5000 iterations with 1000 burn-in samples.

Results in terms of median Effective Sample Size (ESS) and prediction accuracies as Area Under Receiver Operating Characteristic (AUROC) are summarized in Table \ref{tab:BLR_acc}.
All the results are averages over 5 independent runs with random initialization.
In general, SGMGT-D performs better than SGMGT.
For higher-dimensional dataset Cavaran, the performance of $a=2$ decreases significantly, indicating numerical difficulties.
The performance gap between SGMGT and SGMGT-D with $a=1$ or $a=2$ is usually larger than the gap between SGNHT ($a=0.5$).
Presumably when $a$ is greater than 1, SGMGT-D has better convergence.

\begin{figure*}[t!]
	\centering
	\includegraphics[height=31mm,clip,trim=0 0 0 0mm]{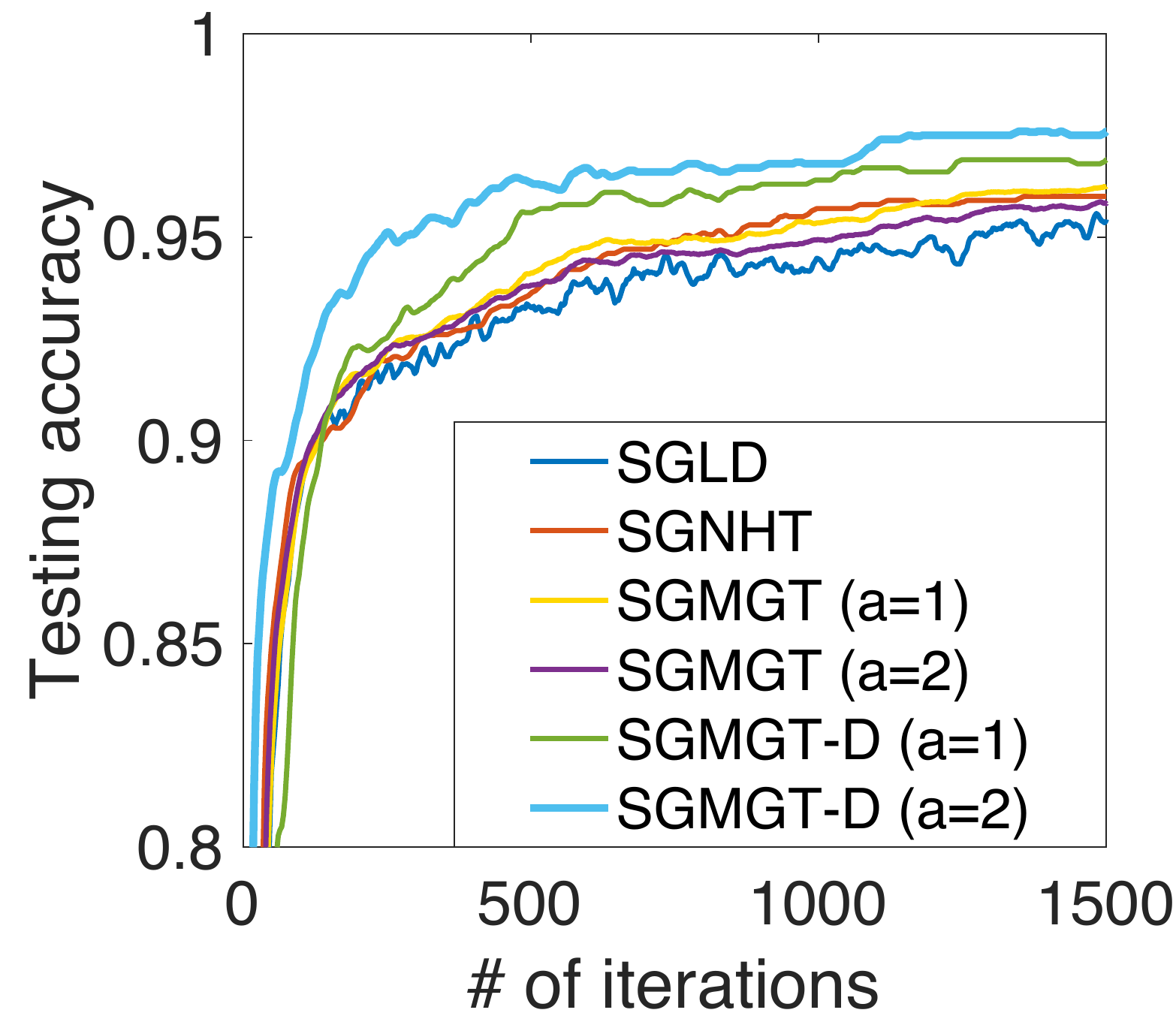}\hspace{5mm}
	\includegraphics[height=31mm,clip,trim=0 0 0 0mm]{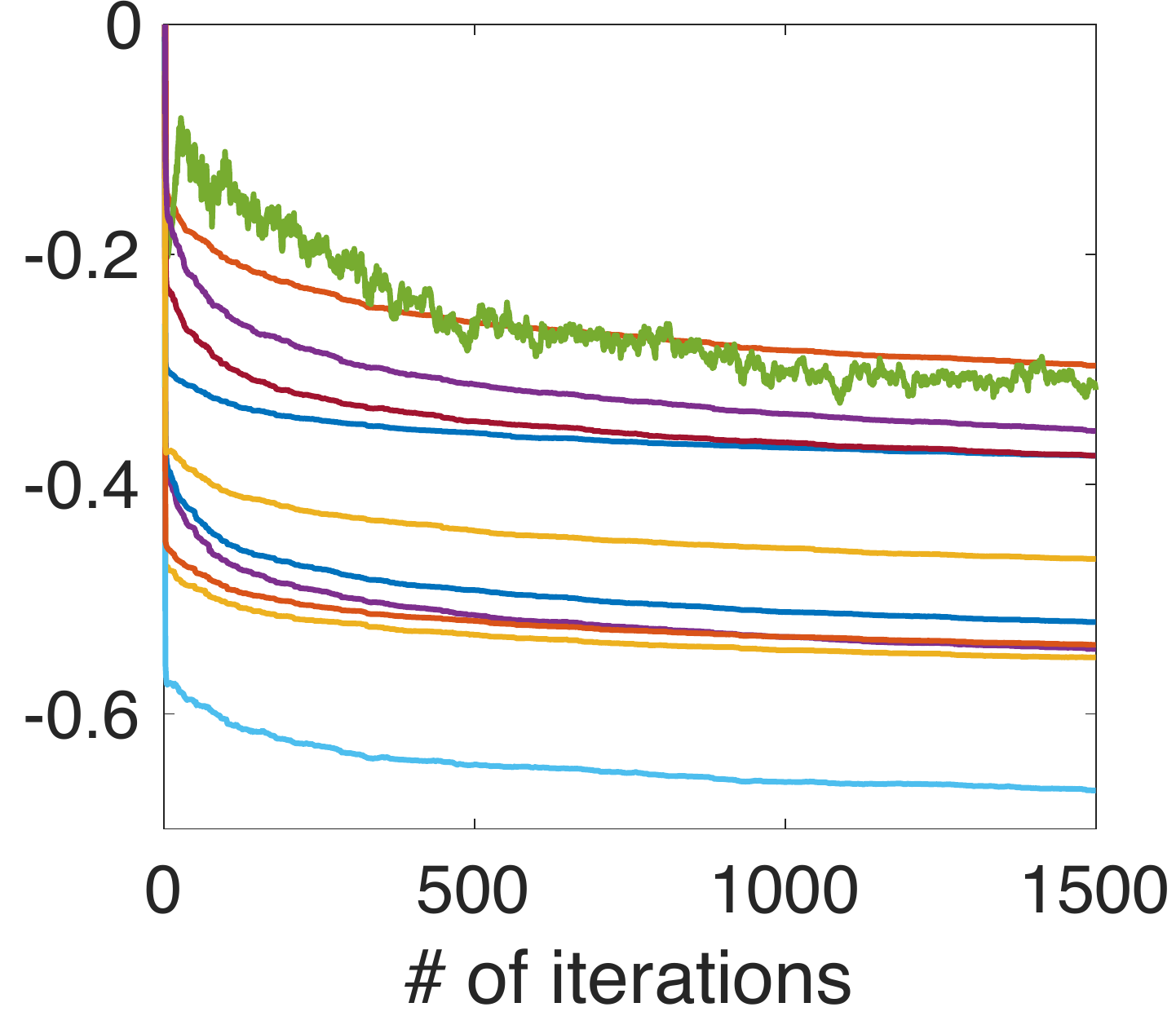}\hspace{5mm}
	\includegraphics[height=31mm,clip,trim=0 0 0 0mm]{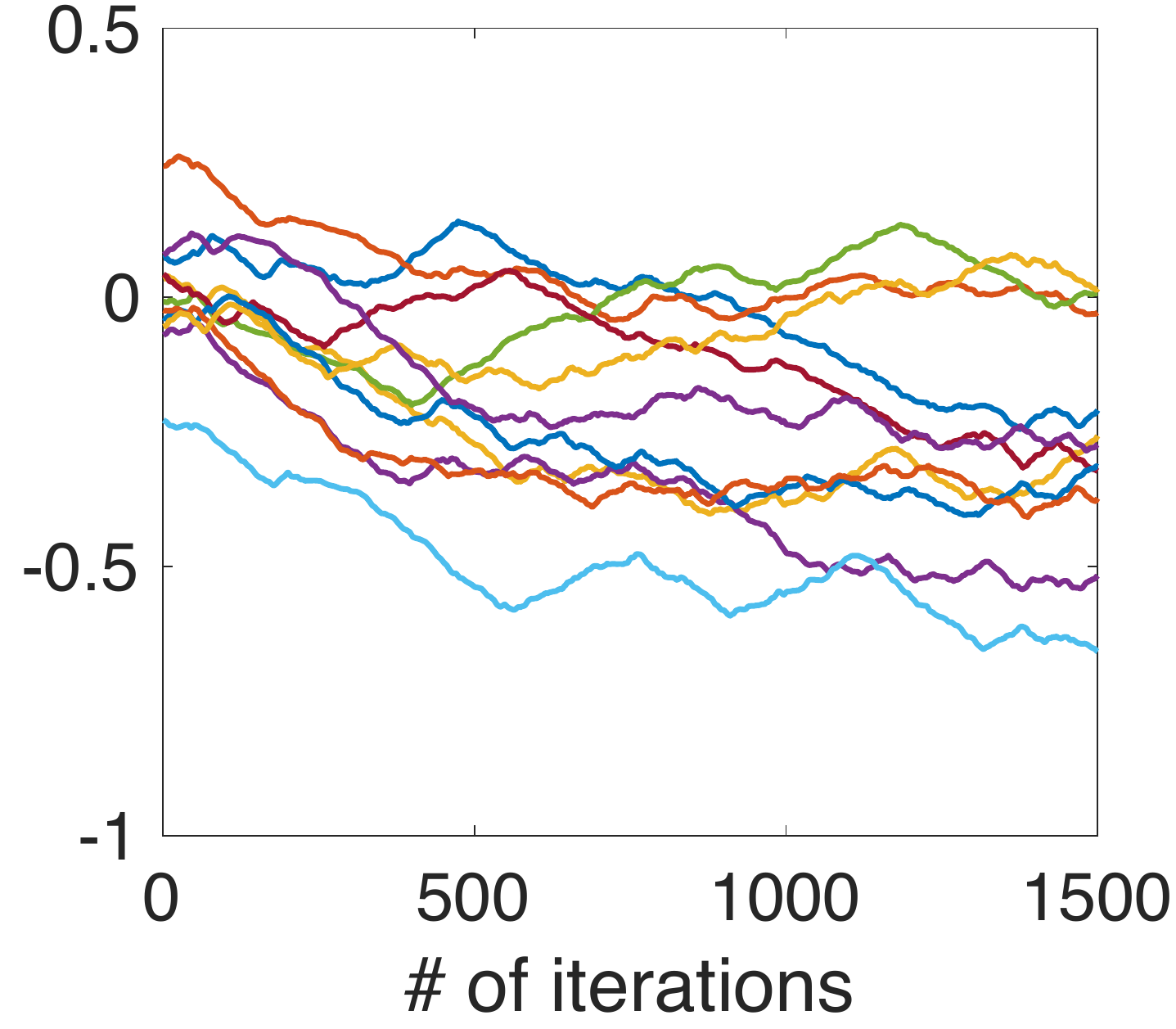}\hspace{5mm}
	\includegraphics[height=31mm,clip,trim=0 0 0 0mm]{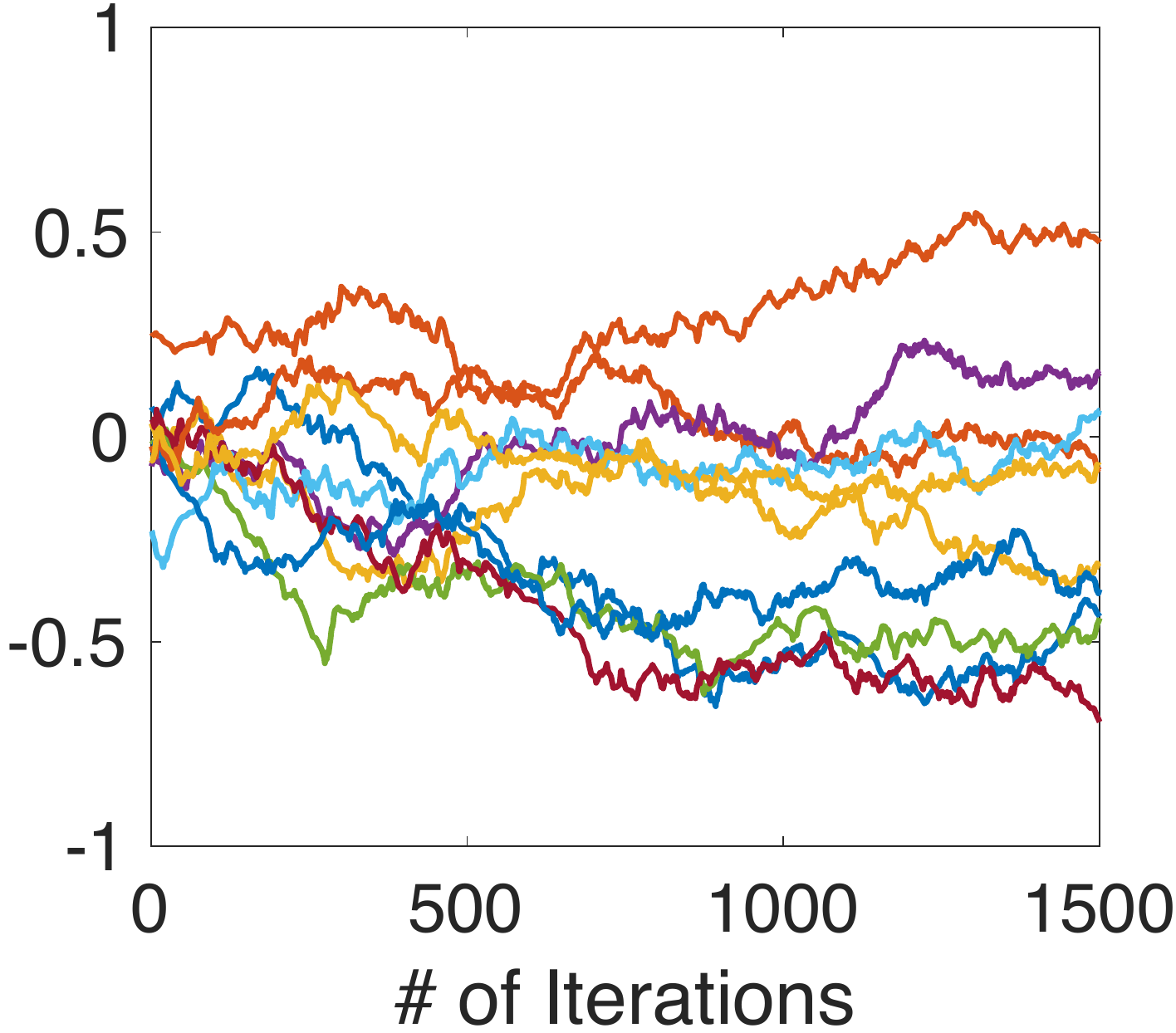}
	\vspace{-2mm}
	\caption{Experimental results for DRBM. Left: testing accuracies for SGLD, SGNHT, SGMGT and SGMGT-D. Middle-left through right: traceplots for SGLD, SGNHT and SGMGT-D with $a=2$, respectively.}
	\label{fig:drbm}
	\vspace{-3mm}
\end{figure*}
\subsection{Latent Dirichlet Allocation}
We also test our methods on Latent Dirichlet Allocation (LDA) \cite{Blei:2003:LDA}.
Details of LDA and our implementation are provided in the SM.
%
We use the ICML dataset \cite{chen2013dependent}, which contains 765 documents corresponding to abstracts of ICML proceedings from 2007 to 2011.
After stopword removal, we obtain a vocabulary size of 1918 and about 44K words.
We use 80\% of the documents for training and the remaining 20\% for testing.
The number of topics is set to 30, resulting in 57,540 parameters.
We use a symmetric Dirichlet prior with concentration $\beta=0.1$.
All experiments are based on 5000 MCMC samples with 1000 burn-in rounds.
We set the minibatch size to 16.
Other hyperparameter settings are provided in the SM.

\begin{table}[t!]
	\caption{The test perplexity with varying stepsize.}
	\label{tab:LDA_ppl}
	\vskip 0.05in
	\begin{center}
		\resizebox{0.48\textwidth}{!}{
		\begin{tabular}{cccccccccc}
		stepsize & 0.04&0.05 &0.06 &0.07 &0.08 &0.09 &  0.1\\
		\hline
		SGLD  & 1058 & 1054 & 1058 & 1067 & {\bf 1037} & 1048 & 1057\\
		SGNHT & 1104 & 1144 & 1039 & {\bf 1024} & 1043 & 1067 & 1107\\
		SGMGT(a=1) & 996 & 988 & 990 & {\bf 986} & 996 & 998 & 997\\
		SGMGT-D(a=1) & 987 & {\bf 983} & 996 & 996 & 992 & 1013 & 1029\\ 
		SGMGT(a=2) & 1024 & 1029 & 1030 & {\bf 1013} & 1030 & 1022 & 1043\\ 
		SGMGT-D(a=2) & 968 & 994 & 973 & 957 & 961 & {\bf 954} & 970\\
		\end{tabular}}
	\end{center}
	\vspace{-6mm}
\end{table}

Table~\ref{tab:LDA_ppl} shows the test perplexities for SGMGT and SGMGT-D for different stepsizes. For each method we highlight the best perplexity.
The SGMGT-D with $a=2$ outperforms other methods, however SGMGT with $a=2$ fails to achieve a comparable result with SGMGT with $a=1$, probably because a good initialization is hard to achieve for a high-dimensional distribution.

\subsection{Discriminative RBM}
We applied our SGMGT to the Discriminative Restricted Boltzmann Machine (DRBM) \citep{LarochelleB:ICML08} on MNIST data.
We choose DRBM instead of RBM because it provides explicit stochastic gradient formulas.

We evaluated our methods empirically and compare them with SGNHT.
We use one hidden layer with 500 units. 
For each method we performed 1500 iterations with 200 burn-in samples.
The minibatch size is set to 100.
Details of the hyperparameter settings for SGMGT and SGMGT-D are provided in the SM.
As shown in Figure~\ref{fig:drbm}(right-most 3 panels), we observe that SGMGT-D with $a=2$ yields a superior mixing performance.
For SGMGT-D with $a=2$, the posterior samples demonstrated both rapid local mixing, and long-range movement.
In contrast, SGLD seems trapped into a local mode after around 500 iterations.

Figure~\ref{fig:drbm}(left) shows that SGMGT-D with $a=2$ delivers the fastest convergence with the highest test accuracy, 0.976.
The SGMGT-D improves over SGMGT, while performance of SGMGT-D seems to increase with a large value of $a$.
We observed that the stochastic resampling played a crucial role for SGMGT, as removing the resampling step resulted in a large drop in testing performance and mixing efficiency.

\subsection{Recurrent Neural Network}
We test our framework on Recurrent Neural Networks (RNNs) for sequence modeling \citep{gan2016scalable}.
%
%
%
We consider two tasks: (\emph{i}) polyphonic music prediction; and (\emph{ii}) word-level language modeling, detailed below.
Additional details of the experiment are provided in the SM.

\paragraph{Polyphonic music prediction}
We use four datasets: Piano-midi.de (Piano), Nottingham (Nott), MuseData (Muse) and JSB chorales (JSB)~\cite{boulanger2012modeling}.
Each of these are represented as a collection of 88-dimensional binary sequences, that span the whole range of piano from A0 to C8. 

We use a one-layer LSTM~\cite{hochreiter1997long} model, and set  the number of hidden units to 200.
The total number of parameters is around 200K.
Each model is trained for 100 epochs.
We perform early stopping, while selecting the stepsize and other hyperparameters by monitoring the performance on validation sets.
Updates are performed using minibatches from one sequence.

\paragraph{Language modeling}
The Penn Treebank (PTB) corpus~\cite{marcus1993building} is used for word-level language modeling.
We adopt the standard split (929K training words, 73K validation words, and 82K test words).
The vocabulary size is 10K.
We train a two-layer LSTM model on this dataset.
The total number of parameters is approximately 6M.
Each LSTM layer contains 200 units.
%
%

\begin{table}[h!]
\caption{Test negative log-likelihood results on polyphonic music datasets and test perplexities on PTB using RNN.}
\label{tab:rnn}
\vskip 0.05in
\begin{center}
\resizebox{0.48\textwidth}{!}{
\begin{tabular}{lccccc}
	Algorithms & Piano & Nott & Muse & JSB & PTB \\
	\hline
	SGLD & 11.37 & 6.07 & 10.83 & 11.25 & 127.47\\
	SGNHT & 9.00 & 4.24 & 7.85 & 9.27 & 131.3\\
	SGMGT (a=1) & 7.90 & 4.35 & 8.42 & 8.67 & 120.6\\
	SGMGT (a=2) & 10.17 & 4.64 & 8.51 & 8.84 & 250.5\\
	SGMGT-D (a=1) & {\bf 7.51} & {\bf 3.33} & 7.11 & 8.46 & 113.8\\
	SGMGT-D (a=2) & 7.53 & 3.35 & {\bf 7.09} & {\bf 8.43} &{\bf 109.0}\\
	\hline
	SGD	& 11.13 & 5.26 & 10.08 & 10.81 & 120.44\\
	RMSprop & 7.70 & 3.48 & 7.22 & 8.52 & 120.45\\
\end{tabular}}
\end{center}
\vspace{-2.5mm}
\end{table}

Results are shown in Table~\ref{tab:rnn}.
The best log-likelihood results on the test set are achieved by using SGMGT-D with either $a=1$ or $a=2$ (depending on the dataset).
To compare with optimization-based methods, we also include results for
SGD \cite{bottou2010large} and RMSprop \cite{tieleman2012lecture}.
A more comprehensive comparison is provided in the SM.

\begin{figure}[h!]
\centering
\includegraphics[width=0.4\linewidth]{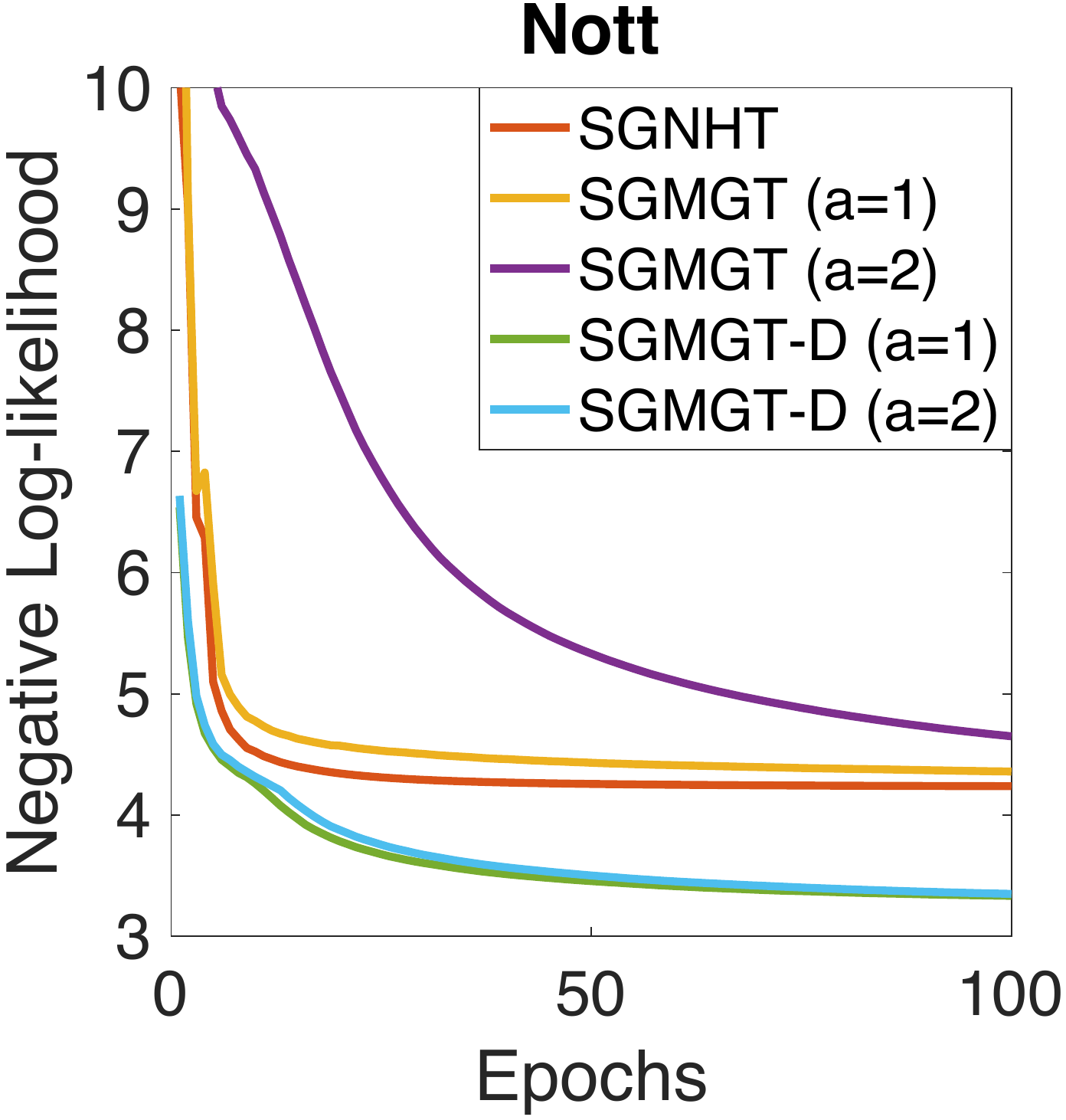}
\includegraphics[width=0.4\linewidth]{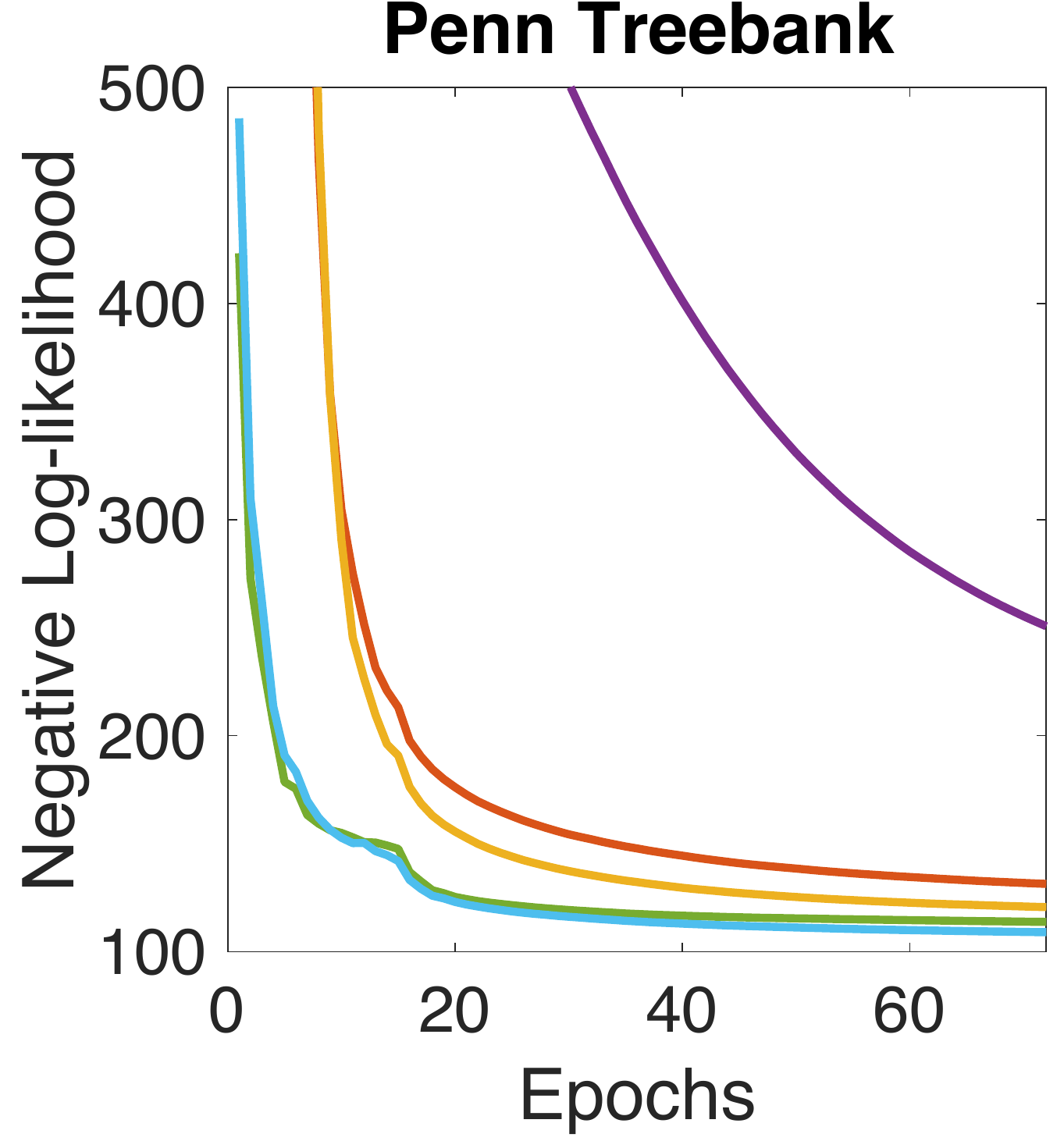}
\vspace{-2mm}
\caption{Learning curves of different SG-MCMC methods on sequence modeling using RNN. Left: Nott. Right: Penn Treebank.}
\label{fig:rnn}
\vspace{-2mm}
\end{figure}

Learning curves for Nott and PTB datasets are shown in Figure~\ref{fig:rnn}.
We omit the SGLD results since they are not comparable with other methods.
For both datasets, we observe that SGMGT-D delivers fastest convergence.
The best negative log-likelihood is achieved by SGMGT-D $a=1$.
The difference between $a=1$ and $a=2$ is small, though SGMGT-D with $a=2$ seems to decrease slightly faster after 20 epochs for PTB data.

We also observe that the SGMGT with $a=2$ seems suboptimal compared with SGMGT with $a=1$ and SGNHT. We hypothesize that numerical difficulties hinder the success of SGMGT with $a=2$, especially in higher-dimensional cases, and without the additional Langevin components of SGMGT-D.





\section{Conclusions}
We improve upon existing SG-MCMC methods with several generalizations.
We employed a general kinetic function, which we have shown to have better mixing efficiency, especially for multimodal distributions.
Since practical use of the generalized kinetics is limited by convergence issues during burn-in,
we injected additional Langevin dynamics and incorporated a stochastic resampling step to obtain generalized SDEs that alleviate the convergence issues.
Possible areas of future research include designing an algorithm in a slice-sampling fashion, which maintains the invariant distribution by leveraging the connections between HMC and slice sampling \citep{zhang2016laplacian}.
In addition, it is desirable to design an algorithm that can adaptively choose the monomial parameter $a$, thereby achieving better mixing while automatically avoiding numerical difficulties.

\section*{Acknowledgments} 
This research was supported by ARO, DARPA, DOE, NGA, ONR and NSF.

\bibliography{sghmc}

\begin{thebibliography}{34}
\providecommand{\natexlab}[1]{#1}
\providecommand{\url}[1]{\texttt{#1}}
\expandafter\ifx\csname urlstyle\endcsname\relax
  \providecommand{\doi}[1]{doi: #1}\else
  \providecommand{\doi}{doi: \begingroup \urlstyle{rm}\Url}\fi

\bibitem[Bache \& Lichman(2013)Bache and Lichman]{bache2013uci}
Bache, Kevin and Lichman, Moshe.
\newblock {UCI} machine learning repository, 2013.

\bibitem[Blei et~al.(2003)Blei, Ng, and Jordan]{Blei:2003:LDA}
Blei, David~M., Ng, Andrew~Y., and Jordan, Michael~I.
\newblock Latent {D}irichlet allocation.
\newblock \emph{JMLR}, 3, 2003.

\bibitem[Bottou(2010)]{bottou2010large}
Bottou, L{\'e}on.
\newblock Large-scale machine learning with stochastic gradient descent.
\newblock In \emph{COMPSTAT}, 2010.

\bibitem[Boulanger-Lewandowski et~al.(2012)Boulanger-Lewandowski, Bengio, and
  Vincent]{boulanger2012modeling}
Boulanger-Lewandowski, Nicolas, Bengio, Yoshua, and Vincent, Pascal.
\newblock Modeling temporal dependencies in high-dimensional sequences:
  Application to polyphonic music generation and transcription.
\newblock In \emph{ICML}, 2012.

\bibitem[Bris \& Lions(2008)Bris and Lions]{bris2008existence}
Bris, C~Le and Lions, P-L.
\newblock Existence and uniqueness of solutions to fokker--planck type
  equations with irregular coefficients.
\newblock \emph{Communications in Partial Differential Equations}, 33\penalty0
  (7):\penalty0 1272--1317, 2008.

\bibitem[Brunick et~al.(2013)Brunick, Shreve, et~al.]{brunick2013mimicking}
Brunick, Gerard, Shreve, Steven, et~al.
\newblock Mimicking an it{\^o} process by a solution of a stochastic
  differential equation.
\newblock \emph{The Annals of Applied Probability}, 23\penalty0 (4):\penalty0
  1584--1628, 2013.

\bibitem[Bubeck et~al.(2015)Bubeck, Eldan, and Lehec]{bubeck2015finite}
Bubeck, Sebastien, Eldan, Ronen, and Lehec, Joseph.
\newblock Finite-time analysis of projected langevin monte carlo.
\newblock In \emph{NIPS}, 2015.

\bibitem[Chen et~al.(2013)Chen, Rao, Buntine, and Whye~Teh]{chen2013dependent}
Chen, Changyou, Rao, Vinayak, Buntine, Wray, and Whye~Teh, Yee.
\newblock Dependent normalized random measures.
\newblock In \emph{ICML}, 2013.

\bibitem[Chen et~al.(2015)Chen, Ding, and Carin]{chen2015convergence}
Chen, Changyou, Ding, Nan, and Carin, Lawrence.
\newblock On the convergence of stochastic gradient mcmc algorithms with
  high-order integrators.
\newblock In \emph{NIPS}, pp.\  2278--2286, 2015.

\bibitem[Chen et~al.(2016)Chen, Carlson, Gan, Li, and Carin]{chen2015bridging}
Chen, Changyou, Carlson, David, Gan, Zhe, Li, Chunyuan, and Carin, Lawrence.
\newblock Bridging the gap between stochastic gradient mcmc and stochastic
  optimization.
\newblock In \emph{AISTATS}, 2016.

\bibitem[Chen et~al.(2014)Chen, Fox, and Guestrin]{chen2014stochastic}
Chen, Tianqi, Fox, Emily~B, and Guestrin, Carlos.
\newblock Stochastic gradient hamiltonian monte carlo.
\newblock \emph{ArXiv}, 2014.

\bibitem[Dalalyan(2016)]{dalalyan2016theoretical}
Dalalyan, Arnak~S.
\newblock Theoretical guarantees for approximate sampling from smooth and
  log-concave densities.
\newblock \emph{Journal of the Royal Statistical Society: Series B (Statistical
  Methodology)}, 2016.

\bibitem[Ding et~al.(2014)Ding, Fang, Babbush, Chen, Skeel, and
  Neven]{DingFBCSN:nips14}
Ding, Nan, Fang, Y, Babbush, R, Chen, C, Skeel, R.~D, and Neven, H.
\newblock Bayesian sampling using stochastic gradient thermostats.
\newblock In \emph{NIPS}, 2014.

\bibitem[Duane et~al.(1987)Duane, Kennedy, Pendleton, and
  Roweth]{duane1987hybrid}
Duane, Simon, Kennedy, Anthony~D, Pendleton, Brian~J, and Roweth, Duncan.
\newblock Hybrid {M}onte {C}arlo.
\newblock \emph{Physics letters B}, 195\penalty0 (2), 1987.

\bibitem[Gan et~al.(2017)Gan, Li, Chen, Pu, Su, and Carin]{gan2016scalable}
Gan, Zhe, Li, Chunyuan, Chen, Changyou, Pu, Yunchen, Su, Qinliang, and Carin,
  Lawrence.
\newblock Scalable bayesian learning of recurrent neural networks for language
  modeling.
\newblock In \emph{ACL}, 2017.

\bibitem[Hochreiter \& Schmidhuber(1997)Hochreiter and
  Schmidhuber]{hochreiter1997long}
Hochreiter, Sepp and Schmidhuber, J{\"u}rgen.
\newblock Long short-term memory.
\newblock \emph{Neural computation}, 1997.

\bibitem[Larochelle \& Bengio(2008)Larochelle and Bengio]{LarochelleB:ICML08}
Larochelle, H. and Bengio, Y.
\newblock Classification using discriminative restricted boltzmann machines.
\newblock In \emph{ICML}, 2008.

\bibitem[Li et~al.(2016)Li, Chen, Fan, and Carin]{li2015high}
Li, Chunyuan, Chen, Changyou, Fan, Kai, and Carin, Lawrence.
\newblock High-order stochastic gradient thermostats for bayesian learning of
  deep models.
\newblock In \emph{AAAI}, 2016.

\bibitem[Lu et~al.(2016)Lu, Perrone, Hasenclever, Teh, and
  Vollmer]{lu2016relativistic}
Lu, Xiaoyu, Perrone, Valerio, Hasenclever, Leonard, Teh, Yee~Whye, and Vollmer,
  Sebastian~J.
\newblock Relativistic monte carlo.
\newblock \emph{arXiv}, 2016.

\bibitem[Ma et~al.(2015)Ma, Chen, and Fox]{ma2015complete}
Ma, Yi-An, Chen, Tianqi, and Fox, Emily.
\newblock A complete recipe for stochastic gradient mcmc.
\newblock In \emph{NIPS}, pp.\  2917--2925, 2015.

\bibitem[Marcus et~al.(1993)Marcus, Marcinkiewicz, and
  Santorini]{marcus1993building}
Marcus, Mitchell~P, Marcinkiewicz, Mary~Ann, and Santorini, Beatrice.
\newblock Building a large annotated corpus of english: The penn treebank.
\newblock \emph{Computational linguistics}, 1993.

\bibitem[Mattingly et~al.(2010)Mattingly, Stuart, and
  Tretyakov]{MattinglyST:JNA10}
Mattingly, J.~C., Stuart, A.~M., and Tretyakov, M.~V.
\newblock Construction of numerical time-average and stationary measures via
  {P}oisson equations.
\newblock \emph{SIAM J. NUMER. ANAL.}, 48\penalty0 (2):\penalty0 552--577,
  2010.

\bibitem[Metropolis et~al.(1953)Metropolis, Rosenbluth, Rosenbluth, Teller, and
  Teller]{metropolis1953equation}
Metropolis, Nicholas, Rosenbluth, Arianna~W, Rosenbluth, Marshall~N, Teller,
  Augusta~H, and Teller, Edward.
\newblock Equation of state calculations by fast computing machines.
\newblock \emph{The journal of chemical physics}, 21\penalty0 (6), 1953.

\bibitem[Neal(2003)]{neal2003slice}
Neal, Radford~M.
\newblock Slice sampling.
\newblock \emph{Annals of statistics}, 2003.

\bibitem[Neal(2011)]{neal2011mcmc}
Neal, Radford~M.
\newblock {MCMC} using {H}amiltonian dynamics.
\newblock \emph{Handbook of Markov Chain Monte Carlo}, 2, 2011.

\bibitem[Risken(1984)]{risken1984fokker}
Risken, Hannes.
\newblock Fokker-planck equation.
\newblock In \emph{The Fokker-Planck Equation}, pp.\  63--95. Springer, 1984.

\bibitem[Rumelhart et~al.(1988)Rumelhart, Hinton, and
  Williams]{rumelhart1988learning}
Rumelhart, David~E, Hinton, Geoffrey~E, and Williams, Ronald~J.
\newblock Learning representations by back-propagating errors.
\newblock \emph{Cognitive modeling}, 5\penalty0 (3):\penalty0 1, 1988.

\bibitem[Teh et~al.(2014)Teh, Thi{\'e}ry, and Vollmer]{teh2014consistency}
Teh, Yee~Whye, Thi{\'e}ry, Alexandre, and Vollmer, Sebastian.
\newblock Consistency and fluctuations for stochastic gradient langevin
  dynamics.
\newblock \emph{ArXiv}, 2014.

\bibitem[Tieleman \& Hinton(2012)Tieleman and Hinton]{tieleman2012lecture}
Tieleman, Tijmen and Hinton, Geoffrey.
\newblock Lecture 6.5-rmsprop: Divide the gradient by a running average of its
  recent magnitude.
\newblock \emph{COURSERA: Neural networks for machine learning}, 2012.

\bibitem[Tuckerman(2010)]{tuckerman2010statistical}
Tuckerman, Mark.
\newblock \emph{Statistical mechanics: theory and molecular simulation}.
\newblock Oxford University Press, 2010.

\bibitem[Vollmer et~al.(2016)Vollmer, Zygalakis, and
  Teh]{vollmer2016exploration}
Vollmer, Sebastian~J, Zygalakis, Konstantinos~C, and Teh, Yee~Whye.
\newblock Exploration of the (non-) asymptotic bias and variance of stochastic
  gradient langevin dynamics.
\newblock \emph{Journal of Machine Learning Research}, 17\penalty0
  (159):\penalty0 1--48, 2016.

\bibitem[Welling \& Teh(2011)Welling and Teh]{welling2011bayesian}
Welling, Max and Teh, Yee~W.
\newblock Bayesian learning via stochastic gradient langevin dynamics.
\newblock In \emph{ICML}, 2011.

\bibitem[Zhang et~al.(2016{\natexlab{a}})Zhang, Chen, Henao, and
  Carin]{zhang2016laplacian}
Zhang, Yizhe, Chen, Changyou, Henao, Ricardo, and Carin, Lawrence.
\newblock Laplacian hamiltonian monte carlo.
\newblock In \emph{Joint European Conference on Machine Learning and Knowledge
  Discovery in Databases}, pp.\  98--114. Springer, 2016{\natexlab{a}}.

\bibitem[Zhang et~al.(2016{\natexlab{b}})Zhang, Wang, Chen, Henao, Fan, and
  Carin]{zhang2016monomial}
Zhang, Yizhe, Wang, Xiangyu, Chen, Changyou, Henao, Ricardo, Fan, Kai, and
  Carin, Lawrence.
\newblock Towards unifying hamiltonian monte carlo and slice sampling.
\newblock In \emph{NIPS}, 2016{\natexlab{b}}.

\end{thebibliography}
\bibliographystyle{icml2017}

\end{document}